\documentclass{article} 
\usepackage[submission]{aaai25}  
\usepackage{times}  
\usepackage{helvet}  
\usepackage{courier}  
\usepackage[hyphens]{url}  
\usepackage{graphicx} 
\usepackage{natbib}  
\usepackage{amsmath}  
\usepackage{amssymb}  
\usepackage{microtype} 
\usepackage{caption} 
\usepackage{microtype}
\usepackage{algorithm}
\usepackage{algorithmic}
\usepackage{booktabs}  
\usepackage{multirow}
\usepackage{booktabs} 
\usepackage{newfloat}
\usepackage{listings}
\DeclareCaptionStyle{ruled}{labelfont=normalfont,labelsep=colon,strut=off} 
\floatstyle{ruled}
\newfloat{listing}{tb}{lst}{}
\floatname{listing}{Listing}
\usepackage{bibentry}
\title{3D-CT-GPT: Generating 3D Radiology Reports through Integration of Large Vision-Language Models}

\author {
    Hao Chen\textsuperscript{\rm 1}\thanks{Co-first authors.},
    Wei Zhao\textsuperscript{\rm 2,3,4}\footnotemark[1],
    Yingli Li\textsuperscript{\rm 1},
    Tianyang Zhong\textsuperscript{\rm 1},
    Yisong Wang\textsuperscript{\rm 2},
    Youlan Shang\textsuperscript{\rm 2},
    Lei Guo\textsuperscript{\rm 1},
    Junwei Han\textsuperscript{\rm 1},
    Tianming Liu\textsuperscript{\rm 5},
    Jun Liu\textsuperscript{\rm 2,3}\thanks{Co-corresponding authors. Second corresponding author.},
    Tuo Zhang\textsuperscript{\rm 1}\thanks{Co-corresponding authors. First corresponding author.},
}
\affiliations {
    \textsuperscript{\rm 1}The School of Automation, Northwestern Polytechnical University, Xi’an 710072, China\\
    \textsuperscript{\rm 2}Department of Radiology, The Second Xiangya Hospital, Central South University, Changsha 410011, China\\
    \textsuperscript{\rm 3}Clinical Research Center for Medical Imaging in Hunan Province, Changsha, China\\
    \textsuperscript{\rm 4}Institute of Biomedical and Health Engineering, Shenzhen Institutes of Advanced Technology, Chinese Academy of Sciences, Shenzhen 518055, China\\
    \textsuperscript{\rm 5}The School of Computing, University of Georgia, Athens, GA 30602, USA\\
    chenhao24@mail.nwpu.edu.cn, wei.zhao@csu.edu.cn, zhangtuo.npu@gmail.com, junliu123@csu.edu.cn
}

\begin{document}

\maketitle 

\begin{abstract}
Medical image analysis is crucial in modern radiological diagnostics, especially given the exponential growth in medical imaging data. The demand for automated report generation systems has become increasingly urgent. While prior research has mainly focused on using machine learning and multimodal language models for 2D medical images, the generation of reports for 3D medical images has been less explored due to data scarcity and computational complexities. This paper introduces 3D-CT-GPT, a Visual Question Answering (VQA)-based medical visual language model specifically designed for generating radiology reports from 3D CT scans, particularly chest CTs. Extensive experiments on both public and private datasets demonstrate that 3D-CT-GPT significantly outperforms existing methods in terms of report accuracy and quality. Although current methods are few, including the partially open-source CT2Rep and the open-source M3D, we ensured fair comparison through appropriate data conversion and evaluation methodologies. Experimental results indicate that 3D-CT-GPT enhances diagnostic accuracy and report coherence, establishing itself as a robust solution for clinical radiology report generation. Future work will focus on expanding the dataset and further optimizing the model to enhance its performance and applicability.
\end{abstract}

\section{Introduction}

In recent decades, the field of radiological imaging has undergone revolutionary changes, making the accurate and efficient interpretation of medical images crucial in modern diagnostics 
\cite{chang2024bootstrapping}. The exponential growth of medical imaging data has placed immense pressure on radiologists, who must not only possess extensive diagnostic expertise but also spend considerable time drafting detailed reports, thereby increasing their workload. Consequently, developing an automated system capable of generating accurate and timely diagnostic reports is essential for alleviating the burden on physicians, improving workflow efficiency, and ensuring diagnostic quality, particularly in resource-constrained environments where high diagnostic accuracy is critical.
\begin{figure}
    \centering
    \includegraphics[width=1\linewidth]{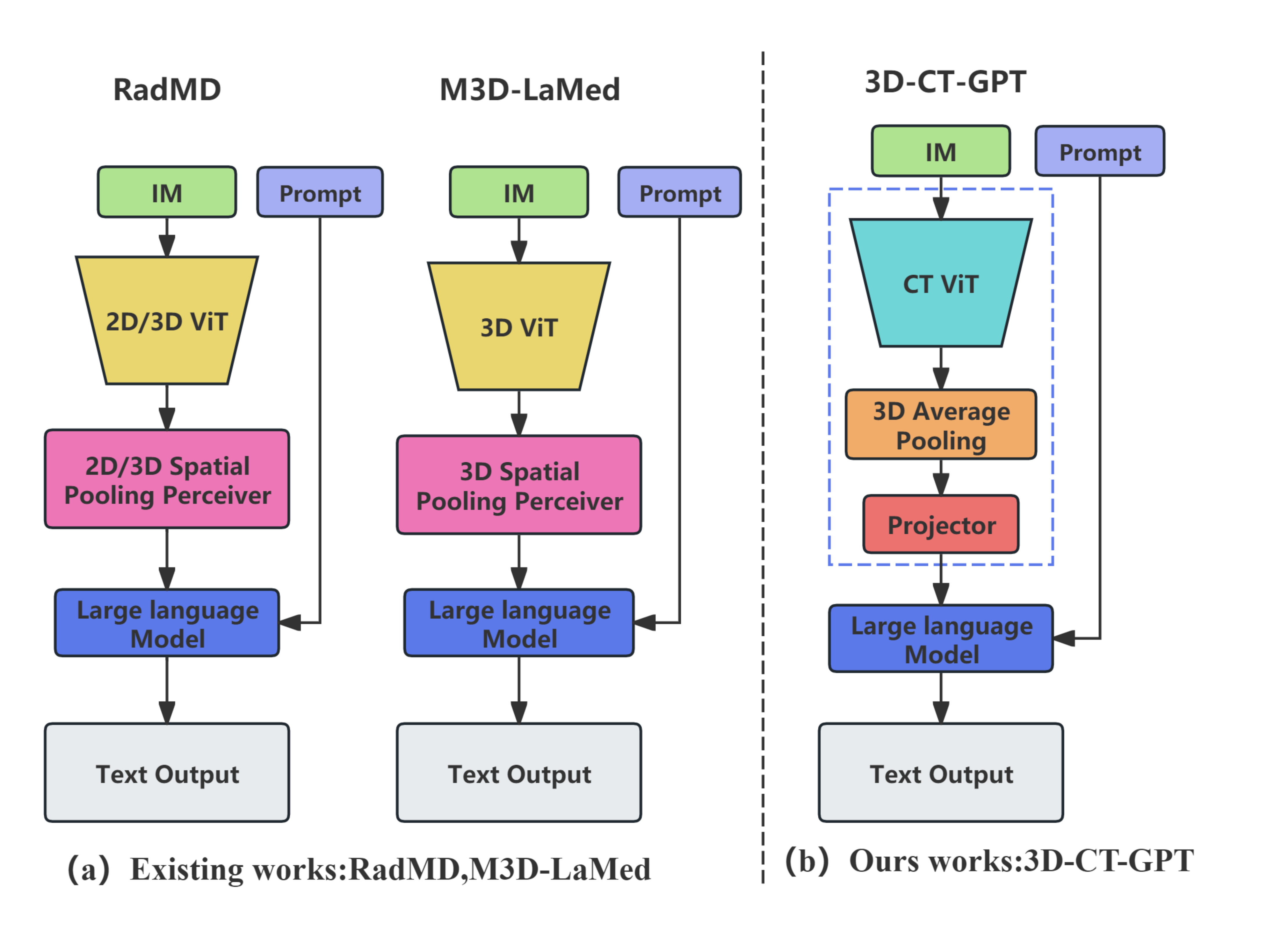}
    \caption{Comparison between: (a) Existing models like RadMD and M3D-LaMed, and (b) Our 3D-CT-GPT, that uniquely combines CT ViT, 3D Average Pooling and a projection layer (dashed box) to enhance report generation from 3D CT scans, improving on past methods.}
    \label{data}
\end{figure}

While significant progress has been made in using machine learning and multimodal language models to automatically generate radiology reports from 2D medical images 
\cite{jing2017automatic,chen2022cross,chen2020generating,qin2022reinforced}, the generation of reports from 3D medical images remains relatively unexplored. This is primarily due to challenges related to data scarcity \cite{li2023systematic}, the complexity of feature extraction, and the computational demands associated with processing large datasets. Three-dimensional imaging modalities, such as Computed Tomography (CT) and Magnetic Resonance Imaging (MRI), offer a more comprehensive view of patient conditions compared to 2D imaging \cite{muller2002computed}, capturing intricate anatomical details and providing higher diagnostic accuracy, especially in detecting complex pathologies that 2D imaging might miss. Despite these advantages, generating high-quality reports from 3D images remains a significant challenge.

Existing methods, such as RadFM \cite{Wu2023TowardsGF}, CT2Rep \cite{hamamci2024ct2rep}, and M3D-LaMed \cite{bai2024m3d}, have initiated exploration into generating radiology reports from 3D images. Although RadFM and M3D-LaMed demonstrate capabilities in tasks like anomaly detection and case retrieval, they fall short when applied to the more intricate task of accurate and coherent report generation from complex 3D medical images. A critical limitation is their insufficient integration of large language models (LLMs) with 3D imaging data, which hampers the generation of diagnostically valuable and contextually accurate reports. CT2Rep, while pioneering in the automated generation of 3D chest CT reports, suffers from an overly complex architecture and high computational demands, limiting its practicality and scalability in real-world clinical settings. Thus, there remains a significant gap in developing a streamlined, efficient, and accurate approach to 3D radiology report generation that effectively harnesses the power of LLMs within an optimized framework.

To address these challenges and advance the development of automatic report generation for 3D medical imaging, we present 3D-CT-GPT, a novel medical visual language model based on Visual Question Answering (VQA) \cite{7410636}, specifically designed for generating radiology reports from 3D CT scans, with a particular focus on chest CTs. As illustrated in Figure \ref{data}, our model uniquely combines CT ViT, 3D Average Pooling, and a projection layer, setting it apart from existing approaches.
Our model's contributions to the field of 3D medical image-based radiology report generation are:
\begin{itemize}
    \item \textbf{CT-Specific Integration of 3D Imaging and Language Models}: We achieve direct and accurate radiology report generation from 3D CT scans by seamlessly integrating CT ViT with a large language model, enhancing both the accuracy and coherence of the generated reports.
    
    \item \textbf{Efficient Training and Data Utilization}: By optimizing our training strategies—pre-training on public datasets followed by fine-tuning on small-scale private datasets—we significantly reduce data requirements while maintaining superior performance.
    
    \item \textbf{Computational Efficiency and Scalability}: Our model is designed for computational efficiency, ensuring scalability and practicality, particularly in resource-constrained environments.
    
    \item \textbf{Robust Generalization and Controlled Text Generation}: The model demonstrates strong generalization across diverse datasets and achieves a balanced trade-off between report diversity and accuracy through an optimized temperature control mechanism.
\end{itemize}

\begin{figure*}[!htb]  
	\centering  
	\includegraphics[width=1\textwidth]{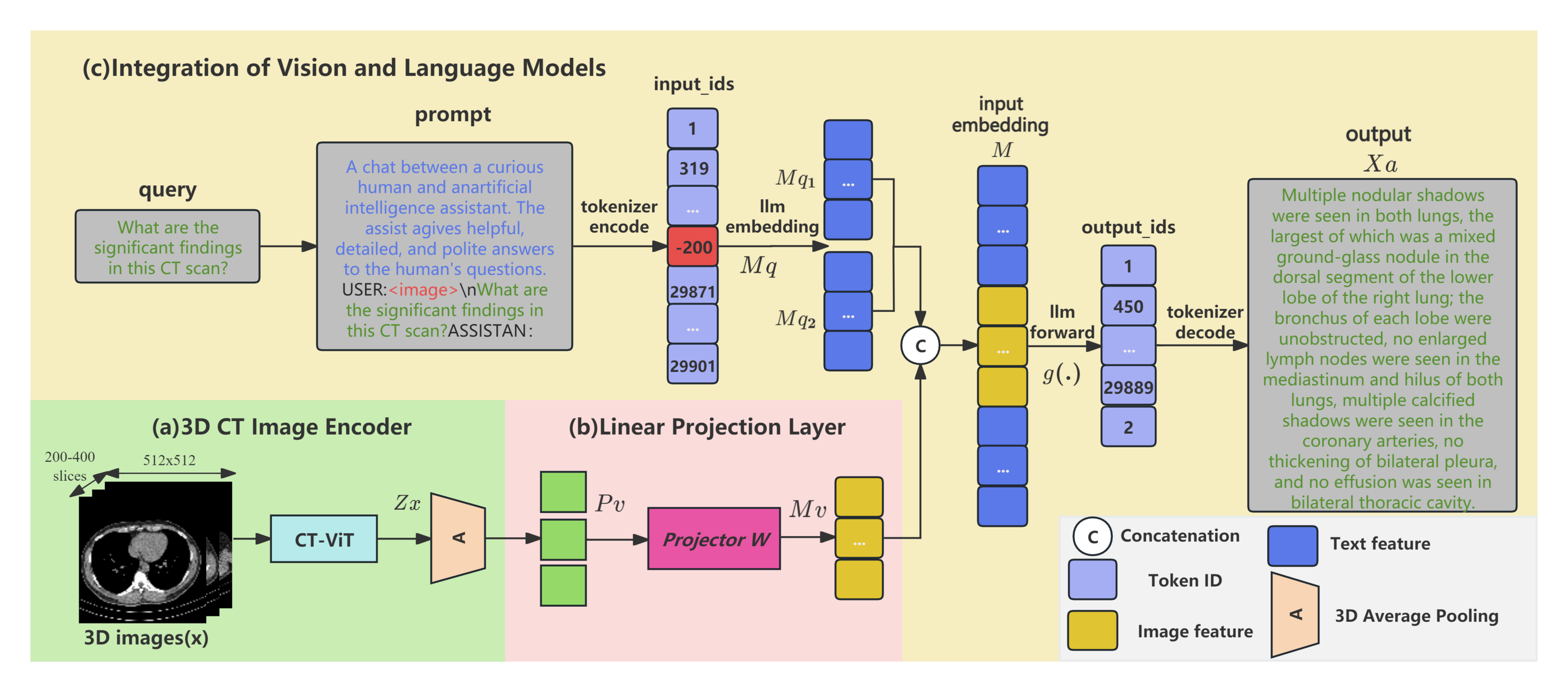}  
	\caption{Overview of the 3D-CT-GPT model architecture, featuring three key components: (a) 3D CT Image Encoder utilizing CT-ViT for feature extraction; (b) Linear Projection Layer for feature transformation; (c) Integration of Vision and Language Models for generating contextually relevant radiology reports.}
	\label{fig1}   
\end{figure*}

\section{Related Works}

\subsection{Large Language Models in Medical Imaging}
The extraordinary generative capabilities of large language models (LLMs) have opened new avenues in natural language processing and computer vision \cite{touvron2023llama,gan2023ziya2,tian2023chimed}. Large Language Vision Models (LLVMs) aim to bridge the gap between visual and textual information, allowing machines to understand and generate content that synthesizes these modalities. Recent research has demonstrated the potential of LLVMs in various tasks, including image captioning \cite{zhu2023minigpt}, visual question answering \cite{bazi2023vision,liu2023q2atransformer,maaz2023video}, and image generation \cite{zhang2023adding}, with applications extending into multimodal scenarios \cite{li2023blip,zhu2023minigpt}, including the medical field \cite{thawkar2023xraygpt,tu2024towards}. As a result, employing LLMs for the automatic generation of medical imaging analysis reports has emerged as a more effective approach. Several existing studies have utilized LLMs for the analysis of 2D medical images, such as X-rayGPT \cite{thawkar2023xraygpt}, LLava-Med \cite{li2023llavamed}, ChatDoctor \cite{li2023chatdoctor}, Med-Alpaca \cite{quispe2023development}, PMC-LLaMA \cite{wu2024pmc}, Clinical Camel \cite{faye2018camel}, DoctorGLM \cite{xiong2023doctorglm}, and Huatuo \cite{wang2023huatuo}. These models are typically initialized with open-source LLMs and fine-tuned on specialized biomedical datasets tailored to specific guidelines. The resulting LLMs hold significant potential, particularly in understanding patient needs and providing informed recommendations.

\subsection{3D Medical Image Analysis and Report Generation}

The field of 3D medical image analysis and report generation has seen significant advancements, yet it continues to face numerous challenges. CT-CLIP \cite{hamamci2024foundation}, utilizing the CT-RATE dataset, has effectively aligned textual information with 3D medical images, enhancing the accuracy of multi-abnormality detection and case retrieval tasks. However, its application in report generation remains limited. RadFM \cite{Wu2023TowardsGF}, a foundational model in radiology, employs large-scale multimodal datasets for pre-training, leading to improvements in report generation accuracy. Nevertheless, it struggles with long sentence generation, 3D image handling, and evaluation metrics. CT2Rep\cite{hamamci2024ct2rep}, the first method specifically designed for generating chest CT reports, leverages advanced 3D vision encoders and multimodal fusion modules. While it shows progress in the precision of report generation, data and computational complexity remain significant obstacles. M3D-LaMed \cite{bai2024m3d} focuses on 3D medical image analysis, utilizing a large-scale 3D multimodal medical dataset. Despite its strong performance, challenges persist in enhancing dataset diversity and managing task complexity. Overall, these methods lay the groundwork for automating 3D medical image report generation, but challenges related to data scarcity, model complexity, and evaluation standards still require further research and resolution.

\section{Method}
\subsection{Model Architecture}
\subsubsection{3D CT Image Encoder:}
As demonstrated in Figure~\ref{fig1}(a),the CT-ViT encoder(\( \Phi_{\text{enc}}^{\text{ct}} \)), derived from CT-CLIP \cite{hamamci2024foundation}, is utilized to extract features from 3D chest CT volumes by dividing them into smaller patches and embedding these into a lower-dimensional latent space. This process results in embedded CT tokens \( Z_x \), which are used for subsequent analysis.

Given a 3D CT image \( x \) with dimensions \( X \in \mathbb{R}^{240 \times 480 \times 480} \), the image is segmented into non-overlapping patches of size \( 15 \times 30 \times 30 \). Each patch is then mapped into a 512-dimensional space \( D \), resulting in a tensor \( \mathbf{Z_x} \) with dimensions \( B \times T \times \frac{H}{p_h} \times \frac{W}{p_w} \times D \), where \( B \) is the batch size, \( T \) represents the number of temporal patches, \( H \) and \( W \) are the height and width of the slices, and \( p_h \), \( p_w \) are the spatial patch dimensions.

The 3D chest CT volume feature extraction process is formally defined as \( Z_x = \Phi_{\text{enc}}^{\text{ct}}(x) \), ensuring the preservation of 3D volumetric information and effectively supporting the construction of sequence-to-sequence models for report generation. To adapt the output for the projection layer, the transformation process for a tensor \( \mathbf{Z_x} \) with dimensions \( B \times T \times \frac{H}{2p_h} \times \frac{W}{2p_w} \times D \) involves the following steps:

First, the tensor is permuted to obtain:
\begin{equation}
\mathbf{Z_x'} = \text{P}(\mathbf{Z_x}, [0, 4, 1, 2, 3])
\end{equation}
resulting in \( \mathbf{Z_x'} \) with dimensions \( B \times D \times T \times \frac{H}{2p_h} \times \frac{W}{2p_w} \).

Next, 3D average pooling is applied with a kernel size of 2, yielding:
\begin{equation}
\mathbf{Z_x''} = \text{A}(\mathbf{Z_x'}, \text{kernel\_size} = 2)
\end{equation}
resulting in \( \mathbf{Z_x''} \) with dimensions \( B \times D \times T \times \frac{H}{2p_h} \times \frac{W}{2p_w} \).

This tensor is then reshaped to:
\begin{equation}
\mathbf{Z_x'''} = \text{R}(\mathbf{Z_x''}, [B, D, T \times \frac{H}{2p_h} \times \frac{W}{2p_w}])
\end{equation}
resulting in \( \mathbf{Z_x'''} \) with dimensions \( B \times D \times (T \times H' \times W') \), where \( H' = \frac{H}{2p_h} \) and \( W' = \frac{W}{2p_w} \).

Finally, a permute operation is applied to obtain:
\begin{equation}
\mathbf{P_v} = \text{P}(\mathbf{Z_x'''}, [0, 2, 1])
\end{equation}
producing \( \mathbf{P_v} \) with dimensions \( B \times (T \times H' \times W') \times D \).

The complete transformation is summarized as:
\begin{equation}
\mathbf{P_v} = \text{P}\left(\text{R}\left(\text{A}\left(\text{P}\left(\Phi_{\text{enc}}^{\text{ct}}(x), [0, 4, 1, 2, 3]\right)\right)\right)\right)
\end{equation}

Here, \( \mathbf{P_v} \) represents the output tensor that contains embedded feature tokens for each batch, preserving crucial 3D volumetric information, which is essential for accurate report generation. yielding the desired output, where each batch contains 512 feature tokens, each of dimension 512.

\subsubsection{Linear Projection Layer:}
For simplicity and efficiency, we opted for a straightforward linear projection technique for input projection, drawing inspiration from the architecture of the LLaVA multimodal large language model. As demonstrated in Figure~\ref{fig1}(b),specifically, we employ a trainable projection matrix \( \mathbf{W} \) to align the CT image token embeddings \( \mathbf{P_v} \) with the semantic space of text word embeddings. This is accomplished by applying the projection matrix to the image token embeddings, resulting in language-aligned embeddings \( \mathbf{M_v} = \mathbf{W} \times \mathbf{P_v} \). This transformation ensures that the dimensionality of the image tokens matches that of the word embeddings used in the language model, thereby enabling efficient and effective data fusion between the visual and textual modalities.

\subsubsection{Integration of Vision and Language Models:}
As illustrated in Figure~\ref{fig1}(c),
the \textit{query} is concatenated with an image placeholder and combined with a dialogue template to form the \textit{prompt}. This \textit{prompt} is then processed by the LLM's tokenizer, which converts the text into tokens. Each token corresponds to a specific token ID based on the LLM’s vocabulary. The image placeholder is assigned a special token ID of -200, serving as a marker within the LLM's vocabulary.

Using the LLM’s embeddings, the text tokens are mapped into word vectors \( \mathbf{M_q} \). The dimensionality of these word vectors varies according to the size of the language model. Since the image placeholder does not have a corresponding entry in the vocabulary, its token embeddings are split into \( \mathbf{M_{q_1}} \) and \( \mathbf{M_{q_2}} \). The image feature vectors, output from the linear projection layer and matched to the same dimensionality, are then concatenated with the split word vectors to form the full input \( \mathbf{M} = \text{concat}([ \mathbf{M_{q_1}}, \mathbf{M_v}, \mathbf{M_{q_2}} ]) \). This combined input is fed into the language model to produce output token IDs, which are then decoded by the LLM’s embedding layer to generate the final output \( \mathbf{X_a} \).

Thus, the LLM function is defined as \( g(\cdot) \). The overall computation can be expressed as:

\begin{equation}
\mathbf{X_a} = g\left(\text{concat}\left( \left[ \mathbf{M_{q_1}}, \mathbf{M_v}, \mathbf{M_{q_2}} \right] \right)\right)
\end{equation}

\subsection{Dataset}

\subsubsection{Data Collection:}
We adopted a subset of a public dataset, \textit{CT-RATE} \cite{hamamci2024foundation}. It includes 25,692 non-contrast chest CT volumes, which have been expanded to 50,188 volumes through various reconstruction techniques, representing 21,304 unique patients, and is further enriched with corresponding radiology text reports, multiple abnormality labels, and metadata. We selected 8,070 cases as the foundational data for our study. Additionally, we collected 2,000 3D chest CT scans and corresponding radiology reports from a renowned international hospital, designated as the private \textit{Dataset-XY}. In this dataset, patients’ ages range from 20 to 88 years, with a mean age of 51.42 years. The gender distribution is 44.7\% female and 55.3\% male.

Each CT volume in \textit{Dataset-XY} has an axial screen resolution of 512x512 pixels, with the number of slices per volume ranging from 100 to 600. Each CT volume in \textit{CT-RATE} has the same axial screen resolution and slice count range as \textit{Dataset-XY}. The \textit{CT-RATE} had already undergone strict anonymization procedures, eliminating the need for further de-identification or format conversion.

\subsubsection{Data Preprocessing:}

For \textit{Dataset-XY}, we first implemented de-identification measures to ensure patient privacy. To maintain the high quality and consistency of the radiology reports with their corresponding 3D chest CT volumes, we conducted rigorous data cleaning, focusing on three key aspects: removing duplicates, correcting data inconsistencies, and filtering out irrelevant text information. Duplicate entries were manually screened to identify and remove redundant elements directly related to data values and report titles, resulting in a standardized and unique dataset. For the image data, we filtered out low-resolution images and eliminated duplicate or irrelevant entries. Then, a meticulous manual review and consolidation process was carried out to ensure dataset uniformity and coherence, ultimately producing a highly optimized dataset consisting of 1,887 cases.

For both datasets, we utilized the slope and intercept values from the metadata to convert CT values to Hounsfield Units (HU), cropping them to the range of [-1000 HU, +200 HU], which reflects the diagnostic limits of the HU scale. Each volume was then resampled to achieve a uniform spacing of 0.75 mm in the x and y axes and 1.5 mm in the z axis, with volumes cropped or padded as necessary to maintain a consistent resolution of 240x480x480. Table \ref{tab:my_label} presents the statistics of the datasets, including the number of radiographic images, the number of reports, and the average report length for the training, testing, and evaluation sets, split at a ratio of 0.8, 0.1, 0.1, respectively.

\begin{table}[h!]
    \centering
    \small  
    \begin{tabular}{l|c c c |c c c}
    \toprule
    \multirow{2}{*}{\textbf{Dataset}} & \multicolumn{3}{|c|}{\textbf{CT-RATE}} & \multicolumn{3}{|c}{\textbf{Dataset-XY}} \\ 
    & \textbf{Train} & \textbf{Test} & \textbf{Val}  & \textbf{Train} & \textbf{Test} & \textbf{Val}  \\
    \midrule
    \textbf{Image} & 6456 & 807 & 807 & 1508 & 190 & 188 \\
    \textbf{Report} & 6456 & 807 & 807 & 1508 & 190 & 188 \\
    \textbf{Avg. Len.} & 198.7 & 196.0 & 198.9 & 88.4 & 88.6 & 88.9 \\
    \bottomrule
    \end{tabular}
    \caption{Table shows the statistics of the two benchmark datasets, including the number of images, reports, and the average word-based length (Avg.Len.) of reports in each set.}
    \label{tab:my_label}
\end{table}

\subsubsection{VQA Dataset Creation:}
To develop a robust Visual Question Answering (VQA) system capable of understanding and generating accurate responses based on 3D medical images, we constructed a specialized dataset that pairs 3D chest CT images with corresponding textual descriptions. This dataset is crucial for training and fine-tuning the model, ensuring that it can accurately interpret complex medical imagery and generate meaningful diagnostic information.

Each entry in the dataset consists of a 3D chest CT image, a related question, and an answer. This structured design enables the model to learn the intricate relationships between visual features in CT scans and their corresponding text, thereby enhancing the system’s comprehension and generation capabilities. During model training, prompts are provided that combine 3D chest CT images with textual information. A specific prompt structure is employed consistently throughout the training process. In this structure, the \(X_{\text{system-message}}\) initiates the interaction, followed by a stopping token \(\texttt{<STOP>}\). The human provides an instruction \(X_{\text{instruct}}\), which is followed by another stopping token. These instructions are randomly selected from a predefined set of prompts, which include various types of queries such as "What findings do you observe in this CT scan?", "Could you summarize the observations from this CT scan?", "What abnormalities are present in this CT scan?", or "How would you interpret the results of this CT scan?". The system then generates a response \(X_a\), in the form of a report corresponding to the 3D chest CT image.

It is noted that, although the VQA system is designed to flexibly handle a wide range of questions related to 3D medical imaging, this paper currently focuses on a specific task—generating radiology reports—to deeply evaluate the core functionality of the model in a controlled environment. Nevertheless, the architecture and dataset have been designed with future scalability in mind, laying the foundation for exploring more complex and diverse medical question-answering tasks in subsequent research.

\begin{table*}[htbp]
\centering
\label{tab:combined_results}
\begin{tabular}{lcccccc}
\toprule
\textbf{Model / Method} & \textbf{BLEU} & \textbf{ROUGE-1} & \textbf{ROUGE-2} & \textbf{ROUGE-L} & \textbf{METEOR} & \textbf{BERTScore\_F1} \\
\midrule
\multicolumn{7}{l}{\textit{\textbf{Training Strategies}}} \\
\midrule
3D-CT-GPT (T1)  & \textbf{0.3836} & \textbf{0.4749} & \textbf{0.2191} & \textbf{0.3281} & \textbf{0.3565} & \textbf{0.8890} \\
3D-CT-GPT (T2)  & \underline{0.3476} & \underline{0.4446} & \underline{0.1978} & \underline{0.3092} & \underline{0.3198} & \underline{0.8862} \\
3D-CT-GPT (T3)  & 0.2323 & 0.3008 & 0.0706 & 0.1567 & 0.2509 & 0.8482 \\
\midrule
\multicolumn{7}{l}{\textit{\textbf{Direct Comparison on Unified Dataset (Unseen by both models)}}} \\
\midrule
3D-CT-GPT (T3) (Private Dataset)  & \textbf{0.2323} & \textbf{0.3008} & \textbf{0.0706} & \textbf{0.1567} & \textbf{0.2509} & \textbf{0.8482} \\
M3D (Private Dataset) & 0.0869 & 0.1336 & 0.0227 & 0.1028 & 0.0710 & 0.8244 \\
\midrule
3D-CT-GPT (T2) (Public Dataset) & \textbf{0.1327} & \textbf{0.2594} & \textbf{0.0586} & \textbf{0.1454} & \textbf{0.1403} & \textbf{0.8412} \\
M3D (Public Dataset)           & 0.0299 & 0.1164 & 0.0223 & 0.0781 & 0.0549 & 0.8203 \\
\midrule
\multicolumn{7}{l}{\textit{\textbf{Indirect Comparison (Literature Results)}}} \\
\midrule
M3D (Linear) (Literature Result)      & 0.1449  & 0.1925  & -      & -      & 0.1411 & 0.8832 \\
M3D (MLP) (Literature Result)         & \underline{0.1515}  & \underline{0.1955}  & -      & -      & \underline{0.1438} & \underline{0.8846} \\
3D-CT-GPT (T1)                        & \textbf{0.3836} & \textbf{0.4749} & 0.2191 & 0.3281 & \textbf{0.3565} & \textbf{0.8890} \\
\midrule
\multicolumn{7}{l}{\textit{\textbf{Ablation Experiments (Based on T2 Strategy)}}} \\
\midrule
3D-CT-GPT (T2-Unfine) & 0.2950 & 0.4163 & 0.1830 & 0.2873 & 0.3037 & 0.8809 \\
3D-CT-GPT (T2)        & \textbf{0.3476} & \underline{0.4446} & \underline{0.1978} & \textbf{0.3092} & \underline{0.3198} & \textbf{0.8862} \\
3D-CT-GPT (T2-Linear) & \underline{0.3418} & \textbf{0.4467} & \textbf{0.1992} & \underline{0.3067} & \textbf{0.3338} & \underline{0.8850} \\
\bottomrule
\end{tabular}
\caption{Performance comparison of 3D-CT-GPT and M3D across different training strategies and datasets. The table presents the evaluation metrics BLEU, ROUGE-1, ROUGE-2, ROUGE-L, METEOR, and BERTScore\_F1 for different models, training strategies, and datasets. The best and second-best results for each metric are highlighted in \textbf{bold} and \underline{underlined}, respectively.}
\end{table*}

\section{Experiment}

Our initial model training process is inspired by conventional multimodal language training methodologies, beginning with pre-training followed by a stage of visual instruction fine-tuning \cite{liu2023improvedllava}.

\subsection{Implementation Details}

\subsubsection{Setup} 
To initialize the model, we leveraged a pre-trained CT-ViT as the visual encoder, alongside the Vicuna-7B model as the large language model (LLM) component. Additionally, to strike an optimal balance between complexity and effectiveness, a randomly initialized linear layer was utilized as the projection module. The training was conducted on a single RTX 3090 GPU (24GB memory). During pre-training, the learning rate was set to 1e-3, and a batch size of 1 was assigned per GPU. For the instruction fine-tuning phase, the learning rate and batch size per GPU were adjusted to 2e-4 and 1, respectively. Both training stages employed the Adam optimizer, a cosine learning rate scheduler, and bfloat16 precision. The pre-training phase required 14GB of GPU memory, while the instruction fine-tuning phase occupied 22GB of GPU memory.

\subsubsection{Stage 1: Pre-training} 
The model aimed to understand the relationship between 3D CT image features and their corresponding reports by analyzing a large collection of 3D CT image-report pairs. During this phase, we froze the image encoder and language model, focusing solely on training the projection layer. The training was conducted using our custom-built VQA dataset. Due to the scarcity of paired 3D CT images and reports, we were unable to employ the large-scale alignment training typical of multimodal models. Instead, we adopted an interactive approach across multiple data types to address this challenge.

\subsubsection{Stage 2: Fine-tuning} 
The model from Stage 1 was further refined to align 3D CT image features with specific radiology reports using image-text pairs. During this phase, we continued to freeze the image encoder but trained both the projection layer and the LLM. Given the constraints on computational resources and the limited dataset, we fine-tuned the language model using a LoRA \cite{hu2022lora} module rather than full parameter tuning. The number of epochs was determined based on avoiding overfitting during training. The high-quality VQA dataset described in Section 4 was again utilized for training. Despite limited resources and sparse datasets, our adequately trained 3D CT-GPT demonstrated the ability to generate more natural and high-quality radiology-specific responses for given 3D chest CT images.

\subsection{Evaluation Metrics}

To evaluate the efficiency of our radiology report generation, we employed natural language generation (NLG) metrics. The primary NLG metrics used include BLEU \cite{papineni-etal-2002-bleu}, METEOR \cite{banerjee-lavie-2005-meteor}, and ROUGE-L \cite{lin-2004-rouge}, which measure word overlap, synonym usage, and sequence matching in the radiology reports, respectively. Specifically, ROUGE scores assess the consistency between the generated text and that of human experts, capturing the presence and order of n-grams, thereby evaluating the quality and coherence of the text. These traditional metrics quantify textual similarity through n-gram overlap or variation, forming a comprehensive framework for assessing automated radiology report generation systems and establishing technical standards for model outputs.

\subsection{Impact of Different Training Strategies on Model Performance}

In the task of medical image generation, the scarcity and large volume of 3D chest CT data pose significant challenges for conducting large-scale model training. To make the most of the limited available datasets and achieve sufficient training, we designed several distinct training strategies and compared the quality of the generated reports. Specifically, our approach utilizes a fine-tuned image encoder and a linear projection module comprising a 2-layer MLP, both adapted with private dataset fine-tuning. To ensure consistent and fair comparison across different training strategies, we selected a uniform temperature parameter (0.7) for all strategies and reported the results based on this setting. The evaluation was conducted on the validation set (\textit{Dataset-XY}\textsubscript{val}), which was not used during the training phase, ensuring that the test results reflect the model’s true generalization performance.

\textit{T1 Training Strategy:} We first pre-trained the model for 5 epochs using \textit{CT-RATE}\textsubscript{train}, followed by fine-tuning with 1508 private data samples (\textit{Dataset-XY}\textsubscript{train}) for 2 epochs. The total time required was 16 hours.

\textit{T2 Training Strategy:} We used 1508 private data samples(\textit{Dataset-XY}\textsubscript{train}) to simultaneously pre-train the model for 5 epochs and fine-tune it for 2 epochs. The total time required was 6 hours.

\textit{T3 Training Strategy:} We pre-trained and fine-tuned the model solely using the public dataset (\textit{CT-RATE}\textsubscript{train}), which took a total of 18 hours.

The results of these strategies are presented in Table 2.

\subsection{Comparison with Existing Methods}

To comprehensively evaluate the performance of 3D-CT-GPT model, we compared it against existing methods, including CT2Rep \cite{hamamci2024ct2rep}, RadFM \cite{Wu2023TowardsGF}, and M3D \cite{bai2024m3d}. Due to significant differences in model architecture and the unavailability of pre-trained weights and inference files for CT2Rep, it was excluded from this comparison. The M3D literature indicates that RadFM does not outperform M3D in generating 3D medical reports, leading us to focus on a fair comparison between our model and M3D. To ensure a balanced evaluation, we adjusted our comparison strategy to account for differences in data processing and model accessibility, employing both direct and indirect assessment methods.

\subsubsection{Direct Comparison on Unified Datasets}

The M3D model processes 3D medical images by stacking multiple 2D PNG images into a 3D format. However, it does not utilize CT values, which are crucial for accurate CT image analysis. Although M3D's pre-trained weights are not publicly available, its complete model architecture and weights are accessible, enabling us to perform inference. To fully assess our model's performance, we converted our data format to be compatible with M3D and used its provided inference code to generate radiology reports.

We first conducted a direct comparison between 3D-CT-GPT (T3 training strategy) and M3D on the private dataset \textit{Dataset-XY}\textsubscript{val}, ensuring that both models were tested under identical conditions. Then, to further evaluate the models' generalization capabilities, we tested 3D-CT-GPT (T2 training strategy) and M3D on the public dataset \textit{CT-RATE}\textsubscript{val}, ensuring the validity and fairness of the comparison.

\subsubsection{Indirect Comparison Using Literature Results}

Since M3D's pre-trained weights weren't publicly available, we couldn't fine-tune the model and instead opted for direct comparison. However, considering that data formatting differences could affect M3D's performance, we acknowledged that direct comparison might introduce some bias. To mitigate this, we used the best evaluation metrics reported in the M3D literature as the benchmark. Simultaneously, we compared these with the performance of 3D-CT-GPT under the T1 training strategy. This approach ensures a fairer assessment of the relative advantages of our 3D-CT-GPT model.

\subsection{Ablation Study}

To assess the impact of different training strategies on model performance, we conducted an ablation study based on the T2 training strategy. The specific experiments are as follows:
\textbf{Comparison 1:}
Using the T2 training strategy, we first fine-tuned the CT-ViT model and compared it with the CT-ViT model without fine-tuning.
\textbf{Comparison 2:}
In the T2 training strategy, we compared the model performance using a 2-layer MLP versus a simple linear projection layer.

\subsection{The Impact of Temperature Parameters on Generation Quality}

The temperature parameter plays a crucial role in controlling the diversity and randomness of the generated text. Lower temperatures make the model's output more deterministic, resulting in more conservative and accurate outcomes, which is particularly suitable for applications where accuracy is paramount, such as medical report generation. In this study, using the T2 training strategy as an example, we adjusted the temperature parameter in our experiments to observe its effect on the quality of the generated reports. To better illustrate the impact of temperature settings on report quality, we show the impact of various temperature settings (ranging from 0.1 to 0.9) on several evaluation metrics, including BLEU, ROUGE, and METEOR in Figure 4. 

\section{Results and Analysis}

\subsection{Comparison of Different Training Strategies}

As shown in Table 2 under the \textit{Training Strategies} section, the performance of the 3D-CT-GPT model varies across different training strategies. The T1 strategy, which combines pre-training on a public dataset with fine-tuning on private data, achieved the highest scores in BLEU (0.3836), ROUGE-1 (0.4749), and METEOR (0.3565), indicating superior report generation quality. However, this approach required the longest training time (16 hours). The T2 strategy, which simultaneously pre-trains and fine-tunes on private data, offered a balanced solution, delivering slightly lower but still competitive performance, making it more suitable for resource-constrained environments. The T3 strategy, trained solely on public data, exhibited lower performance when tested on a private dataset. Nonetheless, it demonstrated the capability to generate reports even in the absence of private data, highlighting the potential of public data in achieving reasonable performance.

\begin{figure}
    \centering
    \includegraphics[width=1\linewidth]{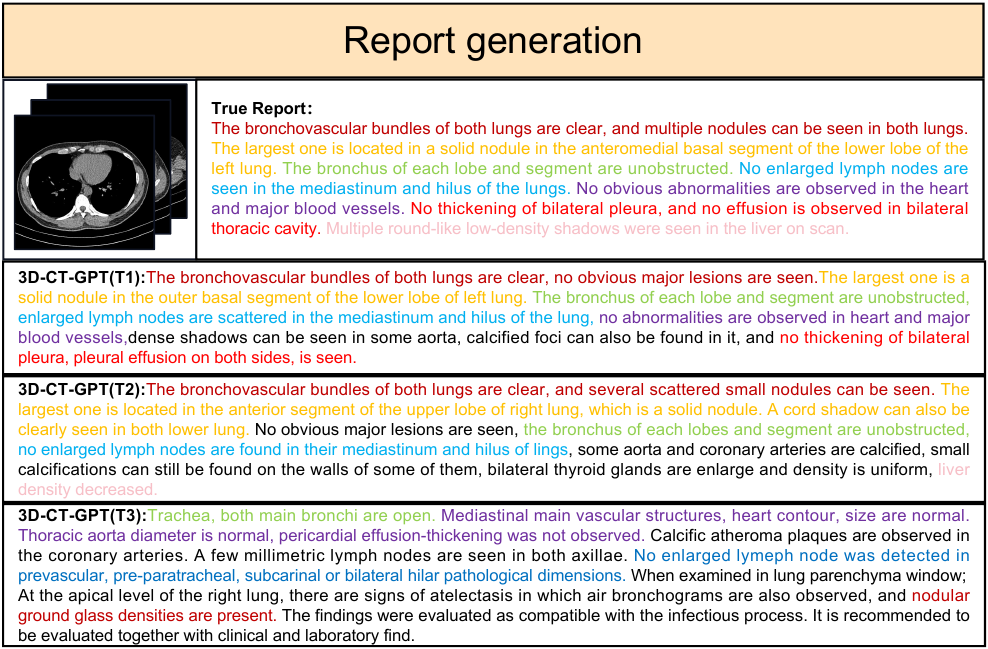}
    \caption{Qualitative comparison of report generation between 3D-CT-GPT and the ground truth. The generated reports from three different training strategies (T1, T2, T3) are compared against the true report. Text highlighted in the same color indicates similar content between the generated reports and the ground truth.}
    \label{fig:enter-label}
\end{figure}

To further illustrate these findings,Figure \ref{fig:enter-label} presents an example where the generated reports from the three different training strategies (T1, T2, and T3) are compared against the ground truth for the same image. It is evident from the comparison that the T1 strategy produces reports that most closely align with the actual diagnostic results.

\subsection{Performance on Unified Dataset}

As shown in the \textit{Direct Comparison on Unified Dataset (Unseen by both models)} section of Table 2, the 3D-CT-GPT model (T3 strategy) significantly outperforms M3D on the private dataset (\textit{Dataset-XY}\textsubscript{val}). Specifically, 3D-CT-GPT achieved higher BLEU (0.2323 vs. 0.0869), ROUGE-1 (0.3008 vs. 0.1336), ROUGE-2 (0.0706 vs. 0.0227), and ROUGE-L (0.1567 vs. 0.1028) scores. 

To evaluate generalization, we tested the T2-trained 3D-CT-GPT on the public dataset (\textit{CT-RATE}\textsubscript{val}), where it continued to outperform M3D, with a BLEU of 0.1327 vs. 0.0299, ROUGE-1 of 0.2594 vs. 0.1164, and ROUGE-L of 0.1454 vs. 0.0781. The METEOR score also favored 3D-CT-GPT (0.1403 vs. 0.0549), highlighting its robustness and clinical potential.

\subsection{Indirect Comparison with Literature Results}
As shown in Table 2 under the \textit{Indirect Comparison (Literature Results)} section, the T1 training strategy outperforms the M3D model in all key metrics, including BLEU (0.3836), ROUGE-1 (0.4749), ROUGE-L (0.3281), METEOR (0.3565), and BERTScore\_F1 (0.8890). In comparison, M3D's reported scores in the literature are lower, with BLEU ranging from 0.1449 to 0.1515, ROUGE-1 from 0.1925 to 0.1955, and METEOR from 0.1411 to 0.1438. These results emphasize the superior capability of 3D-CT-GPT in generating accurate and coherent radiology reports, even without the specific data formatting advantages of M3D. This reinforces the robustness and effectiveness of our model, particularly in leveraging both public and private datasets, and establishes 3D-CT-GPT as a more reliable solution for clinical radiology report generation.
\begin{figure}[h]
    \centering
    \includegraphics[width=1\linewidth]{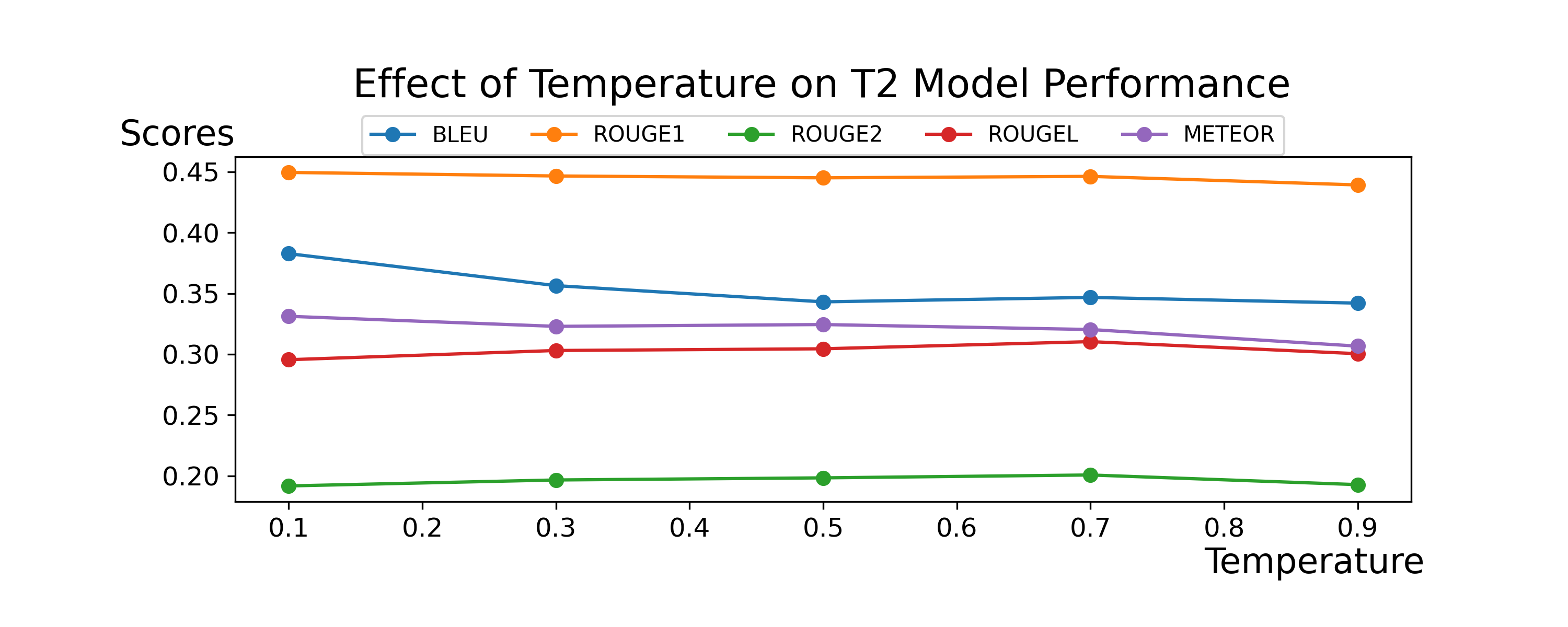}
    \caption{Effect of Temperature on T2 Model Performance across different metrics (BLEU, ROUGE, METEOR).}
    \label{fig:temp_effect}
\end{figure}
\subsection{Ablation Study and Training Strategy Impact}

Under the T2 training strategy, the performance of the 3D-CT-GPT model across three configurations is summarized as follows:

\textbf{Without fine-tuning:} The model showed weaker performance, with BLEU 0.2950, ROUGE-1 0.4163, METEOR 0.3037, and BERTScore\_F1 0.8809.

\textbf{With fine-tuning:} Performance improved significantly, achieving BLEU 0.3476, ROUGE-1 0.4446, METEOR 0.3198, and BERTScore\_F1 0.8862.

\textbf{Using a linear projection layer:} The model maintained strong performance, with BLEU 0.3418, higher ROUGE, METEOR scores, and BERTScore\_F1 0.8850.

Fine-tuning significantly boosts performance, and the linear projection layer offers a slight edge in semantic alignment, providing valuable insights for model optimization.

Higher temperatures increased text diversity but reduced accuracy, as seen in Figure \ref{fig:temp_effect}.

\section{Conclusion}

The experimental results demonstrate that the 3D-CT-GPT model, particularly when employing the T1 and T2 training strategies, significantly outperforms existing methods in generating high-quality radiology reports. The model's strong generalization across diverse datasets, coupled with the benefits of fine-tuning and carefully designed training strategies, underscores its potential for clinical deployment in medical imaging. These findings validate the effectiveness of our approach and open up opportunities for further refinement and broader application in the field of radiology report generation.

\bibliography{aaai25}

\begin{thebibliography}{35}
\providecommand{\natexlab}[1]{#1}

\bibitem[{Antol et~al.(2015)Antol, Agrawal, Lu, Mitchell, Batra, Zitnick, and Parikh}]{7410636}
Antol, S.; Agrawal, A.; Lu, J.; Mitchell, M.; Batra, D.; Zitnick, C.~L.; and Parikh, D. 2015.
\newblock VQA: Visual Question Answering.
\newblock In \emph{2015 IEEE International Conference on Computer Vision (ICCV)}, 2425--2433.

\bibitem[{Bai et~al.(2024)Bai, Du, Huang, Meng, and Zhao}]{bai2024m3d}
Bai, F.; Du, Y.; Huang, T.; Meng, M. Q.~H.; and Zhao, B. 2024.
\newblock M3D: Advancing 3D Medical Image Analysis with Multi-Modal Large Language Models.
\newblock arXiv:2404.00578.

\bibitem[{Banerjee and Lavie(2005)}]{banerjee-lavie-2005-meteor}
Banerjee, S.; and Lavie, A. 2005.
\newblock {METEOR}: An Automatic Metric for {MT} Evaluation with Improved Correlation with Human Judgments.
\newblock In Goldstein, J.; Lavie, A.; Lin, C.-Y.; and Voss, C., eds., \emph{Proceedings of the {ACL} Workshop on Intrinsic and Extrinsic Evaluation Measures for Machine Translation and/or Summarization}, 65--72. Ann Arbor, Michigan: Association for Computational Linguistics.

\bibitem[{Bazi et~al.(2023)Bazi, Rahhal, Bashmal, and Zuair}]{bazi2023vision}
Bazi, Y.; Rahhal, M. M.~A.; Bashmal, L.; and Zuair, M. 2023.
\newblock Vision--language model for visual question answering in medical imagery.
\newblock \emph{Bioengineering}, 10(3): 380.

\bibitem[{Chen et~al.(2022)Chen, Shen, Song, and Wan}]{chen2022cross}
Chen, Z.; Shen, Y.; Song, Y.; and Wan, X. 2022.
\newblock Cross-modal memory networks for radiology report generation.
\newblock \emph{arXiv preprint arXiv:2204.13258}.

\bibitem[{Chen et~al.(2020)Chen, Song, Chang, and Wan}]{chen2020generating}
Chen, Z.; Song, Y.; Chang, T.-H.; and Wan, X. 2020.
\newblock Generating radiology reports via memory-driven transformer.
\newblock \emph{arXiv preprint arXiv:2010.16056}.

\bibitem[{Faye and Bengoumi(2018)}]{faye2018camel}
Faye, B.; and Bengoumi, M. 2018.
\newblock \emph{Camel clinical biochemistry and hematology}.
\newblock Springer.

\bibitem[{Gan et~al.(2023)Gan, Wu, Sun, Lu, Wu, Zhang, Pan, Yang, Yang, Zhang et~al.}]{gan2023ziya2}
Gan, R.; Wu, Z.; Sun, R.; Lu, J.; Wu, X.; Zhang, D.; Pan, K.; Yang, P.; Yang, Q.; Zhang, J.; et~al. 2023.
\newblock Ziya2: Data-centric Learning is All LLMs Need.
\newblock \emph{arXiv preprint arXiv:2311.03301}.

\bibitem[{Hamamci et~al.(2024)Hamamci, Er, Almas, Simsek, Esirgun, Dogan, Dasdelen, Wittmann, Simsar, Simsar et~al.}]{hamamci2024foundation}
Hamamci, I.~E.; Er, S.; Almas, F.; Simsek, A.~G.; Esirgun, S.~N.; Dogan, I.; Dasdelen, M.~F.; Wittmann, B.; Simsar, E.; Simsar, M.; et~al. 2024.
\newblock A foundation model utilizing chest CT volumes and radiology reports for supervised-level zero-shot detection of abnormalities.
\newblock \emph{arXiv preprint arXiv:2403.17834}.

\bibitem[{Hamamci, Er, and Menze(2024)}]{hamamci2024ct2rep}
Hamamci, I.~E.; Er, S.; and Menze, B. 2024.
\newblock Ct2rep: Automated radiology report generation for 3d medical imaging.
\newblock \emph{arXiv preprint arXiv:2403.06801}.

\bibitem[{Hu et~al.(2022)Hu, Shen, Wallis, Allen-Zhu, Li, Wang, Wang, and Chen}]{hu2022lora}
Hu, E.~J.; Shen, Y.; Wallis, P.; Allen-Zhu, Z.; Li, Y.; Wang, S.; Wang, L.; and Chen, W. 2022.
\newblock Lo{RA}: Low-Rank Adaptation of Large Language Models.
\newblock In \emph{International Conference on Learning Representations}.

\bibitem[{Jing, Xie, and Xing(2017)}]{jing2017automatic}
Jing, B.; Xie, P.; and Xing, E. 2017.
\newblock On the automatic generation of medical imaging reports.
\newblock \emph{arXiv preprint arXiv:1711.08195}.

\bibitem[{Li et~al.(2023{\natexlab{a}})Li, Wong, Zhang, Usuyama, Liu, Yang, Naumann, Poon, and Gao}]{li2023llavamed}
Li, C.; Wong, C.; Zhang, S.; Usuyama, N.; Liu, H.; Yang, J.; Naumann, T.; Poon, H.; and Gao, J. 2023{\natexlab{a}}.
\newblock Llava-med: Training a large language-and-vision assistant for biomedicine in one day.
\newblock \emph{arXiv preprint arXiv:2306.00890}.

\bibitem[{Li et~al.(2023{\natexlab{b}})Li, Li, Savarese, and Hoi}]{li2023blip}
Li, J.; Li, D.; Savarese, S.; and Hoi, S. 2023{\natexlab{b}}.
\newblock Blip-2: Bootstrapping language-image pre-training with frozen image encoders and large language models.
\newblock In \emph{International conference on machine learning}, 19730--19742. PMLR.

\bibitem[{Li et~al.(2023{\natexlab{c}})Li, Zhu, Hua, Feng, Bennamoun, Li, Lu, Song, Shen, Xu et~al.}]{li2023systematic}
Li, J.; Zhu, G.; Hua, C.; Feng, M.; Bennamoun, B.; Li, P.; Lu, X.; Song, J.; Shen, P.; Xu, X.; et~al. 2023{\natexlab{c}}.
\newblock A systematic collection of medical image datasets for deep learning.
\newblock \emph{ACM Computing Surveys}, 56(5): 1--51.

\bibitem[{Li et~al.(2023{\natexlab{d}})Li, Li, Zhang, Dan, Jiang, and Zhang}]{li2023chatdoctor}
Li, Y.; Li, Z.; Zhang, K.; Dan, R.; Jiang, S.; and Zhang, Y. 2023{\natexlab{d}}.
\newblock Chatdoctor: A medical chat model fine-tuned on a large language model meta-ai (llama) using medical domain knowledge.
\newblock \emph{Cureus}, 15(6).

\bibitem[{Lin(2004)}]{lin-2004-rouge}
Lin, C.-Y. 2004.
\newblock {ROUGE}: A Package for Automatic Evaluation of Summaries.
\newblock In \emph{Text Summarization Branches Out}, 74--81. Barcelona, Spain: Association for Computational Linguistics.

\bibitem[{Liu et~al.(2024)Liu, Tian, Chen, Song, and Zhang}]{chang2024bootstrapping}
Liu, C.; Tian, Y.; Chen, W.; Song, Y.; and Zhang, Y. 2024.
\newblock Bootstrapping Large Language Models for Radiology Report Generation.
\newblock In Wooldridge, M.~J.; Dy, J.~G.; and Natarajan, S., eds., \emph{AAAI}, 18635--18643.

\bibitem[{Liu et~al.(2023{\natexlab{a}})Liu, Li, Li, and Lee}]{liu2023improvedllava}
Liu, H.; Li, C.; Li, Y.; and Lee, Y.~J. 2023{\natexlab{a}}.
\newblock Improved Baselines with Visual Instruction Tuning.

\bibitem[{Liu et~al.(2023{\natexlab{b}})Liu, Wang, Xu, and Zhou}]{liu2023q2atransformer}
Liu, Y.; Wang, Z.; Xu, D.; and Zhou, L. 2023{\natexlab{b}}.
\newblock Q2atransformer: Improving medical vqa via an answer querying decoder.
\newblock In \emph{International Conference on Information Processing in Medical Imaging}, 445--456. Springer.

\bibitem[{Maaz et~al.(2023)Maaz, Rasheed, Khan, and Khan}]{maaz2023video}
Maaz, M.; Rasheed, H.; Khan, S.; and Khan, F.~S. 2023.
\newblock Video-chatgpt: Towards detailed video understanding via large vision and language models.
\newblock \emph{arXiv preprint arXiv:2306.05424}.

\bibitem[{M{\"u}ller(2002)}]{muller2002computed}
M{\"u}ller, N. 2002.
\newblock Computed tomography and magnetic resonance imaging: past, present and future.
\newblock \emph{European Respiratory Journal}, 19(35 suppl): 3s--12s.

\bibitem[{Papineni et~al.(2002)Papineni, Roukos, Ward, and Zhu}]{papineni-etal-2002-bleu}
Papineni, K.; Roukos, S.; Ward, T.; and Zhu, W.-J. 2002.
\newblock {B}leu: a Method for Automatic Evaluation of Machine Translation.
\newblock In Isabelle, P.; Charniak, E.; and Lin, D., eds., \emph{Proceedings of the 40th Annual Meeting of the Association for Computational Linguistics}, 311--318. Philadelphia, Pennsylvania, USA: Association for Computational Linguistics.

\bibitem[{Qin and Song(2022)}]{qin2022reinforced}
Qin, H.; and Song, Y. 2022.
\newblock Reinforced cross-modal alignment for radiology report generation.
\newblock In \emph{Findings of the Association for Computational Linguistics: ACL 2022}, 448--458.

\bibitem[{Quispe~Bonilla et~al.(2023)Quispe~Bonilla, Quispe~Bonilla, Serrano~Arriezu, Trigo~Vilaseca, Bengoechea, and Quispe~Pe{\~n}a}]{quispe2023development}
Quispe~Bonilla, M.~D.; Quispe~Bonilla, C.; Serrano~Arriezu, L.~J.; Trigo~Vilaseca, J.~D.; Bengoechea, J.; and Quispe~Pe{\~n}a, E. 2023.
\newblock Development and validation of a smart system for medullation and diameter assessment of alpaca, llama and mohair fibres.
\newblock \emph{Animal 17, 5 (2023) 100800}.

\bibitem[{Thawkar et~al.(2023)Thawkar, Shaker, Mullappilly, Cholakkal, Anwer, Khan, Laaksonen, and Khan}]{thawkar2023xraygpt}
Thawkar, O.; Shaker, A.; Mullappilly, S.~S.; Cholakkal, H.; Anwer, R.~M.; Khan, S.; Laaksonen, J.; and Khan, F.~S. 2023.
\newblock Xraygpt: Chest radiographs summarization using medical vision-language models.
\newblock \emph{arXiv preprint arXiv:2306.07971}.

\bibitem[{Tian et~al.(2023)Tian, Gan, Song, Zhang, and Zhang}]{tian2023chimed}
Tian, Y.; Gan, R.; Song, Y.; Zhang, J.; and Zhang, Y. 2023.
\newblock Chimed-gpt: A chinese medical large language model with full training regime and better alignment to human preferences.
\newblock \emph{arXiv preprint arXiv:2311.06025}.

\bibitem[{Touvron et~al.(2023)Touvron, Lavril, Izacard, Martinet, Lachaux, Lacroix, Rozi{\`e}re, Goyal, Hambro, Azhar et~al.}]{touvron2023llama}
Touvron, H.; Lavril, T.; Izacard, G.; Martinet, X.; Lachaux, M.-A.; Lacroix, T.; Rozi{\`e}re, B.; Goyal, N.; Hambro, E.; Azhar, F.; et~al. 2023.
\newblock Llama: Open and efficient foundation language models.
\newblock \emph{arXiv preprint arXiv:2302.13971}.

\bibitem[{Tu et~al.(2024)Tu, Azizi, Driess, Schaekermann, Amin, Chang, Carroll, Lau, Tanno, Ktena et~al.}]{tu2024towards}
Tu, T.; Azizi, S.; Driess, D.; Schaekermann, M.; Amin, M.; Chang, P.-C.; Carroll, A.; Lau, C.; Tanno, R.; Ktena, I.; et~al. 2024.
\newblock Towards generalist biomedical AI.
\newblock \emph{NEJM AI}, 1(3): AIoa2300138.

\bibitem[{Wang et~al.(2023)Wang, Liu, Xi, Qiang, Zhao, Qin, and Liu}]{wang2023huatuo}
Wang, H.; Liu, C.; Xi, N.; Qiang, Z.; Zhao, S.; Qin, B.; and Liu, T. 2023.
\newblock Huatuo: Tuning llama model with chinese medical knowledge.
\newblock \emph{arXiv preprint arXiv:2304.06975}.

\bibitem[{Wu et~al.(2024)Wu, Lin, Zhang, Zhang, Xie, and Wang}]{wu2024pmc}
Wu, C.; Lin, W.; Zhang, X.; Zhang, Y.; Xie, W.; and Wang, Y. 2024.
\newblock PMC-LLaMA: toward building open-source language models for medicine.
\newblock \emph{Journal of the American Medical Informatics Association}, ocae045.

\bibitem[{Wu et~al.(2023)Wu, Zhang, Zhang, Wang, and Xie}]{Wu2023TowardsGF}
Wu, C.; Zhang, X.; Zhang, Y.; Wang, Y.; and Xie, W. 2023.
\newblock Towards Generalist Foundation Model for Radiology.
\newblock \emph{ArXiv}, abs/2308.02463.

\bibitem[{Xiong et~al.(2023)Xiong, Wang, Zhu, Zhao, Liu, Huang, Wang, and Shen}]{xiong2023doctorglm}
Xiong, H.; Wang, S.; Zhu, Y.; Zhao, Z.; Liu, Y.; Huang, L.; Wang, Q.; and Shen, D. 2023.
\newblock Doctorglm: Fine-tuning your chinese doctor is not a herculean task.
\newblock \emph{arXiv preprint arXiv:2304.01097}.

\bibitem[{Zhang, Rao, and Agrawala(2023)}]{zhang2023adding}
Zhang, L.; Rao, A.; and Agrawala, M. 2023.
\newblock Adding conditional control to text-to-image diffusion models.
\newblock In \emph{Proceedings of the IEEE/CVF International Conference on Computer Vision}, 3836--3847.

\bibitem[{Zhu et~al.(2023)Zhu, Chen, Shen, Li, and Elhoseiny}]{zhu2023minigpt}
Zhu, D.; Chen, J.; Shen, X.; Li, X.; and Elhoseiny, M. 2023.
\newblock Minigpt-4: Enhancing vision-language understanding with advanced large language models.
\newblock \emph{arXiv preprint arXiv:2304.10592}.

\end{thebibliography}

\end{document}